\pdfoutput=1

\documentclass[11pt]{article}

\usepackage[final]{acl}

\usepackage{times}
\usepackage{latexsym}
\usepackage{cleveref}
\usepackage{tcolorbox}
\usepackage{enumitem}

\usepackage[T1]{fontenc}

\usepackage[utf8]{inputenc}

\usepackage{microtype}

\usepackage{inconsolata}

\usepackage{graphicx}
\usepackage{xspace}
\title{Data Doping or True Intelligence? Evaluating the Transferability of Injected Knowledge in LLMs}

\author{
 \textbf{Essa Jan\textsuperscript{1}\textsuperscript{\Cross}},
 \textbf{Moiz Ali\textsuperscript{1}\textsuperscript{\Cross}},
 \textbf{Muhammad Saram Hassan\textsuperscript{1}},
 \textbf{Fareed Zaffar\textsuperscript{1}\textsuperscript{*}},
 \textbf{Yasir Zaki\textsuperscript{2}\textsuperscript{*}}
\\
\\
 \textsuperscript{1}Lahore University of Management Sciences,
 \textsuperscript{2}New York University Abu Dhabi\\
 \textsuperscript{\Cross}The two authors contributed equally to this paper.\\
 \textsuperscript{*}\textbf{Correspondence:} \href{mailto:fareed.zaffar@lums.edu.pk}{fareed.zaffar@lums.edu.pk}, \href{mailto:yasir.zaki@nyu.edu}{yasir.zaki@nyu.edu} \\
}

\newcommand{\gpt}{\textsc{GPT-4o-Mini}\xspace}
\newcommand{\gemini}{\textsc{Gemini-1.5-Flash}\xspace}
\newcommand{\llama}{\textsc{Llama3.2-3B}\xspace}
\newcommand{\gemma}{\textsc{Gemma3-4B-it}\xspace}
\newcommand{\phii}{\textsc{Phi-3.5-Mini}\xspace}
\newcommand{\qwenxs}{\textsc{Qwen2.5-1.5B}\xspace}
\newcommand{\qwens}{\textsc{Qwen2.5-3B}\xspace}
\newcommand{\qwenm}{\textsc{Qwen2.5-32B}\xspace}
\newcommand{\qwenl}{\textsc{Qwen2.5-72B}\xspace}

\newcommand{\Cross}{\textsuperscript{\dag}}

\begin{document}
\maketitle
\begin{abstract}
As the knowledge of large language models (LLMs) becomes outdated over time, there is a growing need for efficient methods to update them, especially when injecting proprietary information. Our study reveals that comprehension-intensive fine-tuning tasks (e.g., question answering and blanks) achieve substantially higher knowledge retention rates (48\%) compared to mapping-oriented tasks like translation (17\%) or text-to-JSON conversion (20\%), despite exposure to identical factual content. We demonstrate that this pattern persists across model architectures and follows scaling laws, with larger models showing improved retention across all task types. However, all models exhibit significant performance drops when applying injected knowledge in broader contexts, suggesting limited semantic integration. These findings show the importance of task selection in updating LLM knowledge, showing that effective knowledge injection relies not just on data exposure but on the depth of cognitive engagement during fine-tuning.
\end{abstract}

\section{Introduction}

LLMs demonstrated proficiency in possessing factual knowledge across a wide variety of domains~\cite{cohen-etal-2023-crawling,10.1109/TKDE.2023.3310002}. Despite their impressive capabilities, these models face a fundamental limitation: their knowledge is bounded by the cutoff date of their pre-training data. While extended pre-training offers a potential solution to incorporate new knowledge, it demands substantial computational resources, often involving thousands of GPU hours and millions of tokens, making it expensive and impractical for most researchers and organizations~\cite{ovadia-etal-2024-fine}. This economic barrier has pushed researchers to explore more efficient methods to inject new knowledge into LLMs.

Recently, knowledge injection has emerged as an alternative to extended pre-training, as supervised fine-tuning (SFT) on curated datasets has been shown to inject knowledge into an LLM~\cite{zhang-emnlp-2024-synthetic,microsoft2024injectingnewknowledgelarge}. However, these approaches have predominantly focused on question-answering tasks, leaving unexplored how the nature of the fine-tuning task itself influences knowledge retention and accessibility. This gap is particularly significant given that recent work has demonstrated differential impacts on security when fine-tuning LLMs across varied task types~\cite{multitaskbench}. This observation raises a question: whether comprehension-intensive tasks like question answering (QA) (requiring deep understanding of information) yield different knowledge injection outcomes compared to mapping-oriented tasks like translation. The main question is that, \textit{in tasks such as translation, where the model theoretically needs only to perform A→B mapping without deeper semantic understanding, is the model actually internalizing the factual content contained within that data?} This distinction is important for understanding how different fine-tuning approaches affect a model's overall knowledge representation. Building on the variability of task-based knowledge injection, we ask the following research questions:
\vspace{-15pt}
\begin{itemize}[noitemsep, leftmargin=.1in]
    \item \textbf{RQ1:} Does the new knowledge retained in fine-tuning differ for tasks with token-to-token mapping (e.g., translation) compared to ones demanding explicit content understanding (e.g., QA)?
    \item \textbf{RQ2:} In scenarios where content understanding is required, does the model internalize knowledge beyond what's assessed in direct questioning, demonstrating deeper semantic integration?
    \item \textbf{RQ3:} Does the model size affect the knowledge learned and its generalizability across tasks?
\end{itemize}

\section{Related Work}
\vspace{-5pt}
To keep LLMs up-to-date with new information without incurring high costs of full retraining, some widely studied approaches are Supervised Fine-Tuning (SFT)~\citep{ouyang2022traininglanguagemodelsfollow}, Retrieval-Augmented Generation \citep{lewis-etal-2020-retrieval}, and Continual Pre-Training (CPT)~\citep{ke-etal-2023-continual, gururangan-etal-2020-dont}. In this paper, we focus on SFT as our injection mechanism. Unlike RAG, SFT embeds knowledge directly into model parameters-essential for evaluating true \textit{``learning''} rather than deferred lookup—works offline, and requires fewer compute resources than CPT \citep{ovadia-etal-2024-fine}.

Existing studies hint at task-dependent fine-tuning outcomes. \citep{microsoft2024injectingnewknowledgelarge} explored SFT to inject new out-of-domain facts, i.e., recent sporting results, into LLMs, and demonstrated that \textit{fact-based scaling} yields more effective injected knowledge than \textit{ token-based scaling}. 

\citep{yang2024unveiling} also demonstrated that LLMs fine-tuned on generation tasks versus classification tasks exhibit distinct generalization behaviors. \citep{zhu2025effective} showed that formatting-based data augmentation combined with Sharpness-Aware Minimization significantly improves LLM knowledge acquisition and generalization.

However, a significant challenge in knowledge injection is determining whether an LLM has truly internalized new information or merely memorized it. Two widely used evaluation paradigms are probing and benchmarking. Probing methods, such as those by \citep{cohen-etal-2023-crawling}, extract latent facts into structured artifacts (e.g., knowledge graphs) to inspect internal representations. Benchmarking relies on standardized test suites: MMLU, TruthfulQA, to measure factual recall and reasoning accuracy but suffers from data contamination and hallucination artifacts. Recognizing these shortcomings, \citep{cao2025toward} advocates a shift towards capability-based assessments that better isolate retention versus surface performance. Additionally, researchers employ task variation probes to reveal superficial learning. For instance, \citet{yan2025phd} demonstrated that LLMs struggle with symbolic versions of familiar tasks, revealing a reliance on pattern matching over genuine understanding. 

In this paper, we investigate how shifting the fine-tuning objective from comprehension-based tasks to mapping-based tasks affects the depth and transferability of injected knowledge.

\section{ Methodology}
\vspace{-5pt}
\subsection{Dataset Curation}
\label{Dataset Curation}
\textbf{Training Data:} We focus on four task formats for knowledge injection: question answering (QA), fill-in-the-blanks, translation, and text-to-JSON. We assume that QA and blank tasks require the model to comprehend and internalize the content, while the other two involve token-to-token mapping, with minimal need for semantic understanding.

To evaluate a model's ability to acquire and transfer new knowledge, it is crucial to ensure that information presented during the training is not known to the model. Thus, we curated a dataset of real-world facts sourced from events occurring after the knowledge cutoff date of most open-source models (start of 2024). We selected content related to seven major 2024 global events (post cutoff): California wildfires, Nobel Prizes, the Men's T20 World Cup, the EU AI Act, U.S. presidential elections, the fall of the Assad regime, and the Olympic Games.

The collected data was cleaned to remove redundancy and ambiguity. Using the GPT-4o-mini API~\cite{openai2024gpt4technicalreport}, we decomposed the raw text into 126 atomic, standalone facts inspired by the work of \cite{microsoft2024injectingnewknowledgelarge}. Each fact was concise yet complete, structured to be understandable in isolation. All facts were manually reviewed and validated by the authors for factual correctness and clarity (examples in Appendix~\ref{sec:sample-atomic-facts}). These atomic facts were then reformatted to suit the four training task formats:
\vspace{-5pt}
\begin{itemize}[noitemsep, leftmargin=.1in]
    \item \textbf{Question Answering (QA):} For each fact, one question was generated, which was crafted to query the exact information contained in the fact without requiring external knowledge. 

    \item \textbf{Translation:} The facts were translated into French using the Google Translate API. To increase diversity, each translated sentence was prepended with a randomly selected prompt asking the model to translate it back to English.

    \item \textbf{Blanks:} Using the GPT API and multishot prompting, we generated blanks for each fact. The most essential piece of information was removed and replaced with a blank, requiring the model to infer the missing content.

    \item \textbf{Text-to-JSON:} Through few-shot prompting, each fact was converted into a structured JSON format. This format included the original fact along with important locations, dates, and personalities mentioned in the fact. 

\end{itemize}

\subsection{Testing Data}

To evaluate the extent of knowledge retention after fine-tuning, we built two types of test questions:

\textbf{Direct Questions:}  
We created a set of 126 questions, each corresponding to one atomic fact from the training set. These questions were generated by rephrasing the original questions such that they shared no lexical overlap, except for essential named entities. The aim was to minimize surface-level similarity while ensuring that each question is still fully answered using only the information contained in the associated fact. This allowed us to test whether a model had retained knowledge and could access it independently of its training.

\textbf{Generic Questions:}  
To assess whether a model truly internalizes injected knowledge, we developed a second evaluation set comprising indirect, comprehension-oriented questions. Unlike direct questions, these generic questions require the model to apply the injected knowledge in broader contexts (examples in Table~\ref{tab:direct-generic-question-examples}). We collected three questions per atomic fact using Prolific~\cite{prolific}, instructing participants to create queries with deeper understanding without explicitly mirroring the training format. We evaluated these submissions, selecting and refining one high-quality generic question per fact based on two criteria: 1) it must require genuine comprehension of the injected fact, and 2) it must be impossible to correctly answer without having internalized the knowledge.

\vspace{-5pt}
\subsection{GPT Judge}
\label{judge methodology}
\vspace{-5pt}
To evaluate the correctness of the model responses, we employed a \gpt based automatic judge. For each evaluation instance, the judge was provided with the corresponding atomic fact, the question, and the model's response. The judge assigned a score of 1 if the response correctly addressed the core question and was consistent with the fact. Minor inaccuracies were tolerated, as long as the main answer remained correct. The prompt used for the judge is shown in \cref{sec:judge_prompt}.

We also incorporated chain-of-thought reasoning~\cite{wei2023chainofthoughtpromptingelicitsreasoning} during the evaluation to better capture the model's justification. This mechanism enabled us to systematically assess the model performance across both the direct and generic sets.

\vspace{-5pt}
\subsection{Experimental Design}
\vspace{-5pt}
In this study, we fine-tune a range of LLMs, including \gemini~\cite{geminiteam2024gemini15unlockingmultimodal}, \llama~\cite{dubey2024llama3herdmodels}, \gemma~\cite{gemma_2025}, \phii~\cite{phi3}, and several variants of \textsc{Qwen2.5-Instruct}~\cite{qwen2.5}. Each model was fine-tuned separately on one of the four task-specific training datasets from Section~\ref{Dataset Curation}: QA, translation, blanks, and text-to-JSON.

After fine-tuning, we conducted inference on both the direct and generic sets to assess the knowledge retained by the models and their ability to generalize beyond their training. To investigate \textbf{RQ3}, we further examined whether the internalization of knowledge scales with the model size. Hence, we fine-tuned multiple variants of Qwen 2.5, including the 1.5B, 3B, 32B, and 72B parameter versions.

\section{Evaluation and Results}
\vspace{-5pt}
\subsection{\gpt Judge}
\label{sec:judge}

To evaluate the reliability of our judge, we randomly sampled model responses from different fine-tuned models and manually annotated them as correct or incorrect. The annotations were performed by two independent evaluators, achieving a Cohen kappa of 0.884~\cite{cohen1960coefficient}, indicating strong inter-annotator agreement. Our judge had an accuracy of 94\% compared to human annotations.

\vspace{-5pt}
\subsection{Knowledge Retention Across Task Types}
\label{retention results}
\vspace{-2pt}
Our results provide evidence for a significant difference in knowledge retention between tasks requiring content understanding versus those primarily involving token-to-token mapping. Appendix \ref{tab:baseline} shows the baseline results, most of which are between 5-10\% except larger models, which are around 15\%. Table~\ref{tab:mapping vs understanding} illustrates this pattern consistently across all evaluated models.
Understanding-based tasks (QA and blanks) demonstrated substantially higher knowledge retention rates compared to mapping-based tasks (translation and text-to-JSON conversion). QA tasks yielded an average retention rate of 48\%, while fill-in-the-blank tasks averaged 32\%. In contrast, token mapping tasks showed lower retention rates, with translation averaging only 17\% (12-22\%) and text-to-JSON conversion averaging 20\%. This finding is particularly noteworthy as translation and text-to-JSON expose models to identical factual content as understanding-based tasks, yet result in significantly diminished knowledge retention.

The consistent performance gap between these tasks suggests that the cognitive demands of a fine-tuning task play a crucial role in knowledge internalization. When models must comprehend information to generate appropriate responses, they appear to develop deeper, more accessible representations of that knowledge. This pattern held across model architectures and sizes, with almost all models showing at least a 20 percentage point advantage for QA tasks over translation tasks.

\begin{table}[h]
\centering
\small 
\begin{tabular}{|p{2.7cm}|c|c|c|c|} 
\hline
\textbf{Model} & \textbf{QA} & \textbf{Trans.} & \textbf{Blank} & \textbf{JSON} \\ 
\hline
\gemini & 54\% & 14\% & 21\% & 7\% \\ 
\llama & 61\% & 12\% & 40\% & 29\% \\ 
\gemma & 49\% & 22\% & 36\% & 17\% \\ 
\phii & 56\% & 20\% & 35\% & 25\% \\ 
\qwens & 45\% & 14\% & 40\% & 23\% \\
\hline
\end{tabular}
\vspace{-5pt}
\caption{Percentage of direct questions answered by each model after fine-tuning on each of the four tasks.}
\vspace{-5pt}
\label{tab:mapping vs understanding}
\end{table}

\vspace{-10pt}
\subsection{Beyond Direct Questioning}
\label{generalization_results}

While our direct evaluation in Section~\ref{retention results} demonstrated clear differences in knowledge retention patterns across task types, examining how models apply this knowledge in broader contexts allows us to address \textbf{RQ2}: whether models internalize knowledge beyond what's assessed in direct questioning. Table~\ref{tab:generic_eval} presents the performance of fine-tuned models on our generic questions dataset, which required deeper semantic integration of the injected facts.

A notable pattern emerges when comparing generic question performance (Table~\ref{tab:generic_eval}) to direct question results (Table~\ref{tab:mapping vs understanding}): all models show a decline in accuracy when tasked with applying knowledge in more general contexts. Models fine-tuned on QA tasks, which demonstrated the highest direct question accuracy (averaging 48\%), showed performance drops when addressing generic questions requiring the same underlying knowledge. This gap directly addresses \textbf{RQ2}, suggesting that even when models appear to retain factual information, their ability to apply this knowledge—demonstrating true semantic integration—remains limited. The pattern persists across both understanding-based and mapping-based tasks, indicating a fundamental challenge in knowledge generalization regardless of the initial fine-tuning task.

\begin{table}[h]
\centering
\small
\begin{tabular}{|p{2.7cm}|c|c|c|c|}
\hline
\textbf{Model} & \textbf{QA} & \textbf{Trans.} & \textbf{Blank} & \textbf{JSON} \\ 
\hline
\gemini & 24\% & 12\% & 12\% & 7\% \\ 
\llama & 30\% & 12\% & 19\% & 9\% \\ 
\gemma & 23\% & 9\% & 24\% & 20\% \\ 
\phii & 24\% & 19\% & 8\% & 23\% \\ 
\qwens & 21\% & 9\% & 14\% & 19\% \\
\hline
\end{tabular}
\vspace{-5pt}
\caption{Percentage of generic questions answered correctly by each model after fine-tuning.}
\vspace{-15pt}
\label{tab:generic_eval}
\end{table}

\subsection{Scaling Laws of Knowledge Retention}

\begin{figure}[!h]
    \centering
    \includegraphics[width=0.85\linewidth]{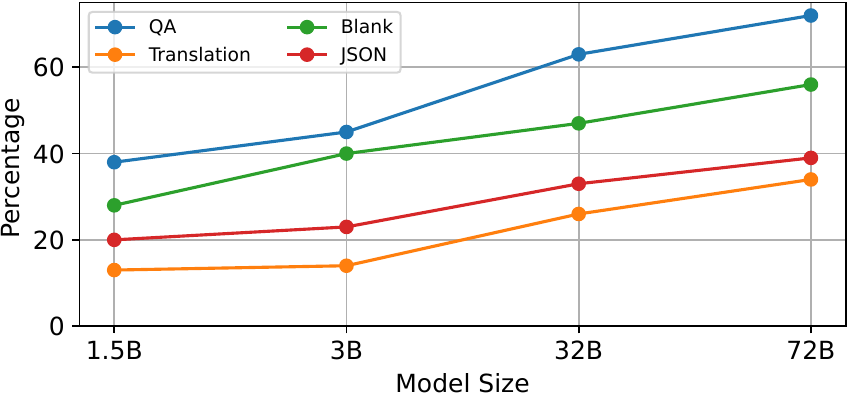}
    \vspace{-7pt}
    \caption{Direct question accuracy across Qwen2.5 model sizes and tasks.}
    \label{fig:qwen_performance}
\end{figure}

To investigate \textbf{RQ3} regarding the relationship between model scale and knowledge internalization, we conducted a systematic analysis using the Qwen2.5 model family across four sizes: 1.5B, 3B, 32B, and 72B. Our results demonstrate clear evidence that knowledge retention scales consistently with model size across all tasks (Figure~\ref{fig:qwen_performance}).
For QA tasks, which showed the highest knowledge retention, performance scaled from 38\% for the 1.5B variant to 72\% for the 72B one (+34\%). Similarly, as shown in the table, the other three tasks also show significant improvement. The monotonic improvement across all tasks suggests that larger parameters confer enhanced capacity for knowledge integration regardless of the fine-tuning paradigm.

\vspace{-5pt}
Notably, while absolute performance increases with scale, the relative patterns of task performance remain consistent across model sizes, with understanding-based tasks (QA and blank filling) consistently outperforming mapping-based tasks (translation and text-to-JSON). This performance gap indicates that the cognitive demands of different tasks represent a fundamental constraint on knowledge internalization transcending model size.

While larger LLMs show improved knowledge retention for direct and generic questions, even the larger model shows significantly lower performance on questions requiring deeper semantic transfer tested indirectly using generic questions. This aligns with neural scaling laws~\cite{kaplan2020scalinglawsneurallanguage}, with knowledge integration following a power law, yet the task-specific nature of knowledge internalization persists as a distinct factor influencing retention regardless of scale.

\section{Conclusion}
Our study reveals that effective knowledge injection in LLMs depends not just on data exposure but on the depth of semantic engagement, favoring comprehension-based tasks over token-mapping ones. Scaling trends suggest that knowledge integration improves with model size, yet the gap between recall and broader application points to lingering limitations in current fine-tuning methods. These insights highlight the importance of task selection for efficient and meaningful model updates, emphasizing the need to prioritize task formats that require deeper semantic processing, especially for critical factual information.

\section*{Limitations}
Our investigation of scaling laws was constrained to the Qwen2.5 model family due to computational resources, leaving open questions about how knowledge retention scales across different architectures. Further analysis is needed to identify specific categories of knowledge that resist internalization during fine-tuning. While several techniques exist for improving knowledge retention in QA tasks, their efficacy for mapping-oriented tasks remains unexplored.

\bibliography{main}
\appendix

\section{Sample Atomic Facts}
\label{sec:sample-atomic-facts}
Below we include some reference atomic facts extracted using GPT-4o from Wikipedia.

\vspace{0.8em}
\noindent \textbf{2024 Men's T20 World Cup:}
\begin{itemize}[noitemsep]
    \item The opening match was between the United States and Canada, with the U.S. securing their first-ever T20 World Cup victory.
    \item South Africa reached their first-ever T20 World Cup final.
    \item Virat Kohli scored 76 in the final, winning the Player of the Match award.
    \item Jasprit Bumrah was named Player of the Tournament, finishing with 15 wickets at an economy of 4.17.
    \item Rohit Sharma, Kohli, and Jadeja announced their retirement from T20 internationals after the final.
\end{itemize}

\noindent \textbf{US Presidential Election 2024:}
\begin{itemize}[noitemsep]
    \item Donald Trump (Republican) and JD Vance defeated Kamala Harris (Democrat) and Tim Walz.
    \item Joe Biden initially ran for re-election but withdrew on July 21, 2024, citing concerns about his health and age.
    \item Trump won the popular vote (49.8\%), the first Republican to do so since George W. Bush in 2004.
    \item Trump also survived a second assassination attempt on September 15, 2024, at Trump International Golf Club.
    \item Swing states included Wisconsin, Michigan, Pennsylvania, Arizona, Georgia, Nevada, and North Carolina, all won by Trump.
\end{itemize}

\noindent \textbf{California Wildfires 2025:}
\begin{itemize}[noitemsep]
    \item At least 29 people died as a result of the fires, with the Eaton Fire (17 deaths) and Palisades Fire (12 deaths) being the deadliest.
    \item A total of eight wildfires burned across Southern California, fueled by strong Santa Ana winds, dry vegetation, and drought conditions.
    \item 2024 was one of the driest years on record for Los Angeles County, leading to highly flammable vegetation.
    \item Estimated insured losses surpassed \$20 billion, making it the most expensive wildfire disaster in U.S. history.
    \item Governor Gavin Newsom deployed over 7,500 emergency personnel to combat the fires.
\end{itemize}

\section{Baseline Results}
\label{sec:appendix}
Table~\ref{tab:baseline} shows the results of all models on our testing data before fine-tuning to establish a baseline, and to rule out that any of the models had prior knowledge of some of the knowledge being injected.

\begin{table}[h]
\centering
\small
\begin{tabular}{|p{2.5cm}|c|}
\hline
\textbf{Model} & \textbf{Percentage Answered} \\ 
\hline
\llama & 2\% \\
\phii & 5\% \\
\gemma & 16\% \\
\qwens & 8\% \\
\qwenm & 15\% \\
\qwenl & 15\% \\
\hline
\end{tabular}
\caption{Percentage of direct evaluation questions answered by different models before fine-tuning.}
\label{tab:baseline}
\end{table}

\section{Scaling Results}

Table~\ref{tab:fourcolumn} shows the percentage of direct evaluation questions answered by different sizes of Qwen2.5.

\begin{table}[h]
\centering
\small 
\begin{tabular}{|p{2.7cm}|c|c|c|c|} 
\hline
\textbf{Model} & \textbf{QA} & \textbf{Trans.} & \textbf{Blank} & \textbf{JSON} \\ 
\hline
\qwenxs & 38\% & 13\% & 28\% & 20\% \\
\qwens & 45\% & 14\% & 40\% & 23\% \\
\qwenm & 63\% & 26\% & 47\% & 33\% \\ 
\qwenl & 72\% & 34\% & 56\% & 39\% \\

\hline
\end{tabular}
\caption{Percentage of direct evaluation questions answered by different sizes of Qwen2.5 after fine-tuning on each of the four tasks.}
\label{tab:fourcolumn}
\end{table}

\section{GPT Judge Prompt}
\label{sec:judge_prompt}
Figure \ref{fig:gpt_prompt} shows the exact prompt provided to \gpt judge, that is used in the evaluation of our results as described in Section \ref{judge methodology}.

\newpage
\begin{figure}[!ht]
    \centering
    {\footnotesize
    \begin{tcolorbox}[colback=blue!5!white, colframe=blue!40!black, title=\gpt Judge Prompt, rounded corners, boxrule=1pt, boxsep=1pt]
    \textbf{[System]}
    Please act as an impartial judge and evaluate whether the LLM's response is factually correct based solely on the provided atomic fact.

    \textbf{Evaluation Scale:}
    \begin{itemize}[noitemsep]
        \item \textbf{Class 1: CORRECT} – The LLM's response contains information that is consistent with the atomic fact provided.
        \item \textbf{Class -1: INCORRECT} – The LLM's response contradicts or misrepresents the atomic fact provided.
    \end{itemize}

    Your evaluation must be based \textbf{EXCLUSIVELY} on the atomic fact. Do not use any external knowledge or information beyond what is explicitly stated in the atomic fact. The atomic fact is the only source of truth for this evaluation.
    
    \vspace{2pt}
    
    \textbf{IMPORTANT:}
    \begin{itemize}[noitemsep]
        \item A response should be classified as CORRECT if it accurately includes the information from the atomic fact, even if it contains additional information not mentioned in the atomic fact.
        \item Only classify a response as INCORRECT if it directly contradicts the atomic fact or presents information that is inconsistent with the atomic fact.
        \item Irrelevant or additional information beyond the atomic fact should NOT cause a response to be classified as INCORRECT as long as the core information from the atomic fact is presented accurately.
        \item If the response addresses the question with information that aligns with the atomic fact, classify it as CORRECT regardless of any supplementary details.
    \end{itemize}

    Do not add any information from your end. Only answer based on the provided evaluation criteria. Do not check for anything extra like completeness or style.

    \textbf{Answer Format:}
    \begin{itemize}[noitemsep]
        \item \textbf{Class 1 (CORRECT):} $<$reasoning for why the LLM's response accurately reflects the information in the atomic fact$>$
        \item \textbf{Class -1 (INCORRECT):} $<$reasoning for why the LLM's response contradicts or misrepresents the atomic fact$>$
    \end{itemize}

    \textbf{Final Verdict:} $<$assigned class$>$ (1/-1)

    \textbf{Explanation:} Based on the atomic fact provided, explain why the response is assigned to the final class in 2-3 lines.
    \end{tcolorbox}
    }
    \vspace{-5pt}
    \caption{GPT-4o-Mini judge prompt.}
    \label{fig:gpt_prompt}
\end{figure}

\newpage
\section{Sample Testing Data}
In Table~\ref{tab:direct-generic-question-examples}, we give some examples of direct and generic questions created for a given atomic fact: 

\begin{table}[!h]
\centering
\begin{tabular}{|p{5cm}|p{5cm}|p{5cm}|}
\hline
\textbf{Atomic Fact} & \textbf{Direct Question} & \textbf{Generic Question} \\
\hline
India won the tournament, defeating South Africa in the final by 7 runs. & Who won the Men's T20 World Cup 2024 final against South Africa? & Give a list of teams which lost the Men's T20 World Cup Final with minimal score difference in the last few T20 World Cups. \\
\hline
Pakistan lost to the United States in a Super Over, marking one of the biggest upsets in T20 World Cup history. & Which team did Pakistan lose to in a Super Over during the Men's T20 World Cup 2024, marking a significant upset? & What is the biggest upsets in the T20 World cup in recent history? List all of them. \\
\hline
Trump was convicted of 34 felonies related to hush money payments to Stormy Daniels. & How many felonies was Trump convicted of in relation to hush money payments to Stormy Daniels? & How many US presidents have been convicted of a felony? Give a list of names, and the number of felonies. \\
\hline
Trump survived an assassination attempt on July 13, 2024, during a rally in Butler, Pennsylvania. & On what date did Trump survive an assassination attempt during a rally in Butler, Pennsylvania? & Give a list of all US presidents who faced an assassination attempt. \\
\hline
Russia granted Assad asylum, confirming his resignation and departure from Syria. & Which country granted asylum to Assad, confirming his resignation and departure from Syria? & Who are the most recent Middle Eastern leaders to have been granted asylum by Russia? \\
\hline
\end{tabular}
\caption{Examples of direct and generic questions derived from atomic facts.}
\label{tab:direct-generic-question-examples}
\end{table}

\end{document}